%% file: main.tex
\documentclass[10pt,twocolumn,letterpaper]{article}

\usepackage{graphicx}
\usepackage{multirow}
\usepackage[pagenumbers]{cvpr} % To force page numbers, e.g. for an arXiv version


\definecolor{cvprblue}{rgb}{0.21,0.49,0.74}
\usepackage[pagebackref,breaklinks,colorlinks,allcolors=cvprblue]{hyperref}

\title{SpikeBottleNet: Spike-Driven Feature Compression Architecture for Edge-Cloud Co-Inference}

\author{Maruf Hassan\\
Centre for Sustainable Digital Technologies\\
Technological University Dublin, Ireland\\
{\tt\small maruf.hassan@tudublin.ie}
\and
Steven Davy\\
Centre for Sustainable Digital Technologies\\
Technological University Dublin, Ireland\\
{\tt\small steven.davy@tudublin.ie}
}

\begin{document}
\maketitle
\input{sec/0_abstract}    
\input{sec/1_intro}
\input{sec/2_background}
\input{sec/3_methodology}
\input{sec/4_evaluation}
\input{sec/5_conclusion}
{
    \small
    \bibliographystyle{IEEEtran}
    \bibliography{main}
}

\input{sec/X_suppl}

\end{document}

%% file: sec/0_abstract.tex
\begin{abstract}
Edge-cloud co-inference enables efficient deep neural network (DNN) deployment by splitting the architecture between an edge device and cloud server, crucial for resource-constraint edge devices. This approach requires balancing on-device computations and communication costs, often achieved through compressed intermediate feature transmission. Conventional DNN architectures require continuous data processing and floating point activations, leading to considerable energy consumption and increased feature sizes, thus raising transmission costs. This challenge motivates exploring binary, event-driven activations using spiking neural networks (SNNs), known for their extreme energy efficiency. In this research, we propose SpikeBottleNet, a novel architecture for edge-cloud co-inference systems that integrates a spiking neuron model to significantly reduce energy consumption on edge devices. A key innovation of our study is an intermediate feature compression technique tailored for SNNs for efficient feature transmission. This technique leverages a split computing approach to strategically place encoder-decoder bottleneck units within complex deep architectures like ResNet and MobileNet. Experimental results demonstrate that SpikeBottleNet achieves up to 256x bit compression in the final convolutional layer of ResNet, with minimal accuracy loss (0.16\%). Additionally, our approach enhances edge device energy efficiency by up to 144x compared to the baseline BottleNet, making it ideal for resource-limited edge devices.
\end{abstract}

%% file: sec/1_intro.tex
\section{Introduction}
\label{sec:intro}

Recently, adopting Deep Neural Networks (DNNs) in mobile and Internet of Things (IoT) devices has driven breakthroughs across a wide array of intelligent applications \cite{DurmazIncel2023}. However, DNN-based applications require substantial computational resources, and their direct implementation incurs significant power and computation costs. The most common deployment approach involves transmitting raw data to a cloud service for inference. While this approach is useful, it leads to considerable communication overhead and latency. Furthermore, the computational power of IoT devices goes unused when data is entirely offloaded to servers. For applications involving personal data, such as health records, local processing and anonymization are required.

An alternative strategy is on-device inference, which entails the deployment of compressed DNNs on mobile devices. Nonetheless, this approach often results in compromised performance due to over-compression. These challenges have prompted the exploration of other solutions, with device-edge co-inference emerging as a viable option \citep{Li2023}. This method partitions a network, enabling a mobile device access to the leading part and an edge server to handle the rest of the computations. By processing the earlier layers on a mobile device and then transmitting the compressed features to the cloud, both latency and energy consumption associated with cloud data transmission can be substantially decreased.
%-------------------------------------------------------------------------

Prior research on device-edge co-inference has primarily focused on traditional deep neural networks, which rely on continuous values for activations \citep{Eshratifar2019}\citep{Shao2020}. Although these networks deliver high performance, they are characterized by considerable computational complexity and energy demands. In contrast, distinguished as the third generation of neural network models, the brain-inspired Spiking Neural Networks (SNNs) offer a promising alternative that is extremely energy efficient when deployed in specialized neuromorphic hardware \citep{Schuman2022}\citep{chandarana2022energy}. The adoption of binary spike signals enables SNNs to utilize low-power sparse additions as opposed to the high-power matrix multiplication operations prevalent in traditional models.

SNNs can inherently support feature compression through sparse representations and temporal coding, thereby reducing the communication overhead. The combination of comparable accuracy with traditional DNN \citep{Guo2023}, reduced energy consumption \citep{su2023deep}\citep{kim2020spiking}\citep{qiu2024gated}, and a high feature compression ratio makes SNNs an optimal choice for application in device-edge co-inference contexts.
%-------------------------------------------------------------------------

This study introduces SpikeBottleNet, an SNN-based architecture for energy-efficient feature compression in device-edge co-inference image classification. The key contributions of this paper are highlighted as follows:

\begin{itemize}
    \item We propose a novel architecture for spike-driven feature compression in device-edge co-inference systems, which achieve comparable performance with traditional DNN based methods.
    \item The experiments on CIFAR100 datasets demonstrate that SpikeBottleNet reduces the ResNet50 and MobileNetV1 intermediate feature sizes by factors of 256x and 174x, respectively, in the final convolutional layer. The model experiences only minimal accuracy degradation of approximately 0.16\% and 0.72\% respectively. 
    \item Our model's performance is achieved using an architecture similar to DNN based BottleNet\citep{Shao2020}, but with a significant reduction in mobile energy consumption, by a factor of 12x on traditional hardware and up to 144x on neuromorphic hardware.
\end{itemize}

The paper proceeds as follows: Section 2 provides a review of related research. Section 3 covers the necessary background and details the SpikeBottleNet architecture. Section 4 describes the experimental setup and presents the results. Finally, Section 5 summarizes key findings and suggests directions for future research.
%-------------------------------------------------------------------------

%% file: sec/2_background.tex
\section{Background}
\label{sec:background}

%-------------------------------------------------------------------------
\subsection{Deep Spiking Residual Network}

Residual Networks (ResNet) address the degradation problem in deep learning, particularly in very deep networks. \citep{he2016deep}. He et al. introduced identity mappings between layers through shortcut connections. This technique significantly facilitated the training of very deep neural networks. Inspired by this breakthrough, Zheng et al. \citep{zheng2021going} established pioneering work in SNN-oriented batch normalization (BN) and demonstrated the successful training of spiking ResNet-34/50 architectures using surrogate gradients within the STBP framework \citep{wu2018spatio}. Extending this progression, Fang et al. \citep{fang2021deep} introduced the spike-element-wise (SEW) ResNet. Furthermore, Hu et al.'s Membrane Shortcut MS-ResNet introduced a novel shortcut for neuron layer membrane potentials, facilitating the training of up to 482-layer network \citep{hu2021advancing}. Beyond ResNet, architectures like MobileNet have been employed in spiking neural networks for energy-efficient, competitive object detection in real-world automotive event data \citep{cordone2022object}. In our study, we chose ResNet50 for its optimal balance of depth and established benchmarks, alongside a lightweight MobileNetV1, to showcase SpikeBottleNet's general applicability.

SNN based models are well-suited for neuromorphic chips, like NorthPole \citep{modha2023neural}, TrueNorth \citep{akopyan2015truenorth}, Loihi \citep{orchard2021efficient}, and ROLLS \citep{indiveri2015neuromorphic}. Unlike traditional architectures constantly processing data, these chips only activate specific neurons when receiving input spikes. This sparse operation significantly reduces power consumption, as a large portion of the chip remains idle. Compared to DNNs, the SNNs running on neuromorphic platforms can achieve up to 1000x greater power efficiency \citep{pei2019towards}. Our proposed SpikeBottleNet architecture aligns with the spike-driven paradigm, making it well-suited for implementation on neuromorphic chips.
%-------------------------------------------------------------------------
\subsection{Model Splitting}
Split Computing (SC) is a technique that segments a DNN model into two parts: the initial segment executes on an edge device, while the subsequent portion operates on a server \cite{matsubara2022split}. This division reduces computational load on the edge device and lowers data transmission costs. Processing delays on an edge server are typically less significant than those encountered with local processing and data transmission delays \citep{matsubara2021neural}. Recent research on dividing models has introduced a bottleneck feature in vision models, which results in a smaller data size than the original input sample \citep{Li2023}\citep{Eshratifar2019}\citep{Shao2020}\citep{matsubara2019distilled}. However, these studies have not leveraged the potential of SNNs in reducing data size through model splitting and bottleneck layer, thereby overlooking an opportunity to further optimize performance.

\subsection{Feature Compression}
The interaction between network splitting and feature compression presents a significant challenge in the device-edge co-inference. The ideal splitting strategy seeks to minimize both on-device communication latency and feature transmission size. Various studies have explored splitting and compression techniques, such as JALAD's use of quantization and Huffman coding for feature compression \citep{li2018jalad}, and a deep feature compression structure with encoder-decoder pairs for edge-cloud collaborative inference \citep{choi2018near}. BottleNet \citep{Eshratifar2019} introduced a bottleneck unit with a trainable lossy compressor, and BottleNet++ optimized feature compression with joint source-channel coding \citep{Shao2020}. However, BottleNet++ neglects on-device model compression, leading to excessive on-device computations at most split points. Recent work introduced a compression module using channel attention and entropy encoding \citep{Li2023}. Existing research has focused only on DNNs, overlooking SNNs, which naturally enable sparse data representation and lower communication overhead—ideal for efficient feature compression
%-------------------------------------------------------------------------

%% file: sec/3_methodology.tex
\begin{figure*}[!ht]
\centering
\includegraphics[width=6in]{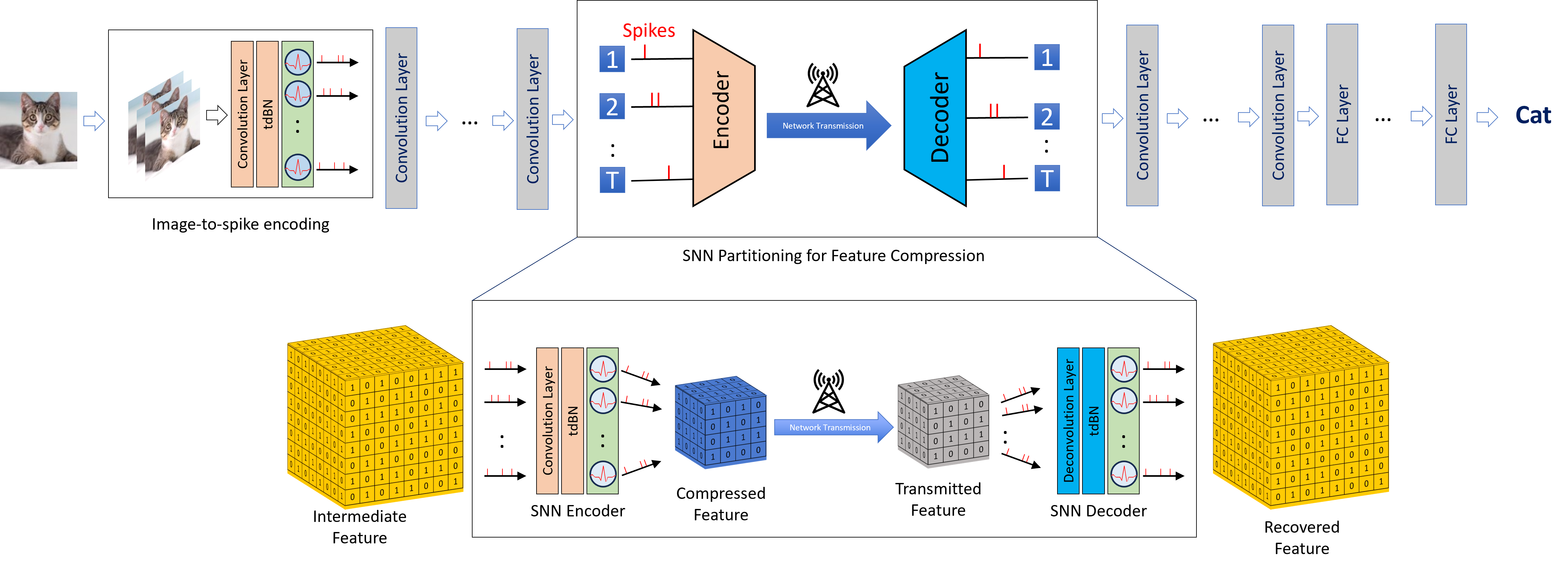}
\caption{Overview of the SpikeBottleNet architecture showing a spiking network with feature compression using an encoder-decoder module. The network partitions intermediate spike tensors at specific layers to optimize communication and energy efficiency in device-edge co-inference.}
\label{fig:spikebottlenet-architecture}
\end{figure*}

\section{Methodology}

In this section, we present the network structure for our proposed model, describing the neuron model, methods for input encoding, SNN partitioning, and feature reduction using a compression layer module, alongside our approach to training the model.

\subsection{Overall Network Architecture}

The architecture of our proposed model primarily comprises the spike residual network and the bottleneck layer positioned at different network split points to efficiently compress and transmit the intermediary feature tensors. As depicted in Figure \ref{fig:spikebottlenet-architecture}, the feature compression module comprises lightweight CNN-based encoder and a decoder. These layers perform convolution operations on spikes produced by Leaky Integrate-and-Fire (LIF) neurons. This setup differs from that of BottleNet by substituting ReLU activation layer with LIF neurons for spike-based processing. Due to the temporal dynamic of SNNs, we use threshold-dependent batch normalization (tdBN) as suggested by Zheng et al. \citep{zheng2021going} to avoid gradient issues and regulate firing rates.

The SNN ResNet architecture uses Residual Blocks (RBs) as its fundamental building units. Each RB is a sequence of operations: {CONV-tdBN-LIF}, repeated three times. An initial encoding layer first converts the input image into spike signals, which are processed through the RBs. Similar to DNN-ResNet, the number of channels doubles when the feature map size is halved. The architecture ends with average pooling and a fully connected layer to aggregate temporal information.

In contrast, the SNN MobileNet architecture leverages Inverted Residual Blocks (IRBs) which consists of depthwise separable convolution blocks (DSCs) to reduce computational complexity. Each IRB performs a depthwise convolution followed by a pointwise convolution. The architecture begins with a standard convolution and uses depthwise separable convolutions in subsequent layers. Similar to ResNet, final stage includes average pooling and a fully connected layer for classification. LIF neurons are placed after each RB in ResNet and IRB in MobileNet, as splitting points are chosen after these blocks to ensure they generate spikes for efficient feature compression and transmission.

\subsection{Spiking Neuron Model}
Spiking neural networks are composed of neurons, which transform synaptic inputs into action potential outputs. In contrast to DNNs, which ignore temporal dynamics and focuses on spatial information transmission, SNNs employ spiking neurons. These neurons simulate the behaviour of biological plausibility through the dynamics of membrane potentials and spike-based communication. The Leaky Integrate-and-Fire (LIF) model is commonly used in SNNs because it offers an optimal balance between biological plausibility and computation complexity alongside reduced energy consumption compared to DNN neurons. Our study adopts the iterative LIF model by Wu et al., applied in SNNs, which uses the Euler method to convert the LIF differential equation into a recursive formula \citep{wu2019direct}:
\begin{equation}
V_t=\tau_{decay}V_{t-1}\ +\ I_t
  \label{eq:lif-equation1}
\end{equation}
where $\tau_{decay}$ is the membrane potential decay rate, $V_t$ represents the membrane potential, and $I_t$ signifies the synaptic input at time $t$. Upon reaching a threshold $V_{th}$, a neuron fires (and $V_t$ resets to 0) when $V_t>V_{th}$. This model integrates spatial structure by considering synaptic inputs as $x_t=\sum_{j}{w_jo_t\left(j\right)}$ where $w_j$ are the synaptic weights and $o_t\left(j\right)$ is the binary spiking output at time $t$. The model is encapsulated by the following equations \citep{zheng2021going}:
\begin{equation}
V_{t,n+1}=\tau_{decay}V_{t-1,n+1}\left(1-o_{t-1,n+1}\right)+x_{t,n} \\
\label{eq:lif-equation2}
\end{equation}

\begin{equation}
o_{t, n+1}= \begin{cases}1, & \text { if } V_{t, n+1}>V_{t h} \\
0, & \text { otherwise }\end{cases}
\label{eq:lif-equation2}
\end{equation}
where $V_{t,n}$ is the neuron’s membrane potential at the $n$-th layer and time $t$, and $o_{t,n}$ represents the binary spike output.

\subsection{Threshold-dependent batch normalization}
SNNs demand a tailored normalization technique due to their unique temporal dimension and activation dynamics. We adopt the TDBN \citep{zheng2021going} normalization method, which operate across both spatial and temporal domains. It focuses on updating the membrane potential $u_i^t$ of a neuron at time $t$, using the TDBN to normalize a mini batch of the sequential inputs $\{I_i^t=\sum_{j=1}^{n}{w_{ij}o_j^t\ |\ t=1,\ \ldots,\ T}\}$ as follows:
\begin{equation}
u_i^t=\tau_{mem}u_i^{t-1}+TDBN\ \left(I_i^t\right)
\label{eq:lif-tdbn1}
\end{equation}
where $TDBN$ operation on input $I_i^t$ is defined as:
\begin{equation}
TDBN\ \left(I_i^t\right)=\lambda_i\frac{\alpha V_{th}(I_i^t-\mu_{c_i})}{\sqrt{\sigma_{c_i}^2+\epsilon}}+\beta_i
\label{eq:lif-tdbn2}
\end{equation}
Here, $\mu_{c_i}$ and $\sigma_{c_i}^2$ are the mean and variance for channel $i$, $\epsilon$ is a small constant for numerical stability, and $\lambda_i$, $\beta_i$, and $\alpha$ are parameters that adjust normalization, with $\alpha$ scaling inputs based on the neuron threshold $V_{th}$.

\subsection{Spike Encoding}
Since SNNs process time-varying data, we convert the static CIFAR100 dataset into spike sequences by passing the same training sample $X\ \in\mathbb{R}^{m\times n}$ repeatedly at each timestep, like transforming the CIFAR100 dataset into a static video. Each data element is normalized to a precision value ranging between 0 and 1: $X_{ij}\in[0,\ 1]$. The initial convolutional layer facilitates the conversion of static images into spike sequences, as illustrated in the Figure \ref{fig:spikebottlenet-architecture}.

\subsection{Bottleneck Unit}
The bottleneck unit architecture significantly enhances data transmission efficiency between edge devices and the server, utilizing a feature reduction unit (encoder) and feature recovery unit (decoder). Both the encoder and decoder integrate three layers: a convolutional layer (or deconvolutional layer for the decoder), a threshold-dependent batch normalization (tdBN) layer, and an activation layer using LIF neurons. The encoder reduces the spatial and channel dimensions. In contrast, the decoder restores the feature dimensions. The tensor dimension changes in this process, starting from $(time\_step, batch\_size, c, w, h)$ and compressing to $(time\_step, batch\_size, c^\prime,\ w^\prime,\ h^\prime)$. This involves narrowing the $c$ channels to $c^\prime$ and adjusting the spatial dimensions to $w^\prime$ and $h^\prime$, which is accomplished by choosing appropriate filter sizes for $c^\prime$ and adjusting convolution kernel strides to $\left\lfloor\frac{w}{w^\prime}\right\rfloor$ and $\left\lfloor\frac{h}{h^\prime}\right\rfloor$. This streamlined architecture minimizes transmission load while maintaining the integrity and usability of the data, making it particularly effective for applications requiring real-time data exchange between edge devices and cloud services.

\subsection{Training Strategy}
In this research, we introduce a novel architecture trainable in an end-to-end fashion. However, direct training of the entire architecture faces slow convergence. Therefore, we adopt a two-step training approach: initially training the SNNs to achieve optimal accuracy \citep{yifan_hu__2024}. The block structures of ResNet or MobileNet provides inherent breakpoints within the network that can be exploited in device edge co-inference scenarios. Next step, we deploy the bottleneck layer at these splitting points. In this step, we adopt the compression-aware training approach from \citep{Eshratifar2019}. The whole model is fine-tuned simultaneously by training and updating the compression module's weights and the SNN's parameters.

Our methodology involves placing the bottleneck unit after certain blocks chosen from the underlying network, at up to $M$ potential splitting points, with $M$ being no greater than the network's $N$ total blocks. For each point, we explore bottleneck layers that implement varying degrees of compression along channel or spatial dimensions. From those models that maintain an accuracy reduction within 2\%, we select the model that requires the highest bit compression ratio. This selection process is replicated across all $M$ locations. Ultimately, we pick the best way to partition the network that minimizes edge device energy consumption and/or maximizes the bit compression ratio, in alignment with our optimization goals.

One challenge in training SNNs is the use of spike functions that can't be differentiated. This hinders the traditional backpropagation process because the gradients can't transmit back through the preceding layers. To overcome this, we employ a surrogate gradient approach \citep{wu2018spatio}, defined as:
\begin{equation}
\frac{\partial X_i^{t,n}}{\partial V_i^{t,n}}=\frac{1}{a}sign(\left|V_i^{t,n}-V_{th}\right|\le\ \frac{a}{2})
\label{eq:surrogate-gradient}
\end{equation}
where $a$ ensures the total gradient adds up to 1 and adjusts the curve's steepness. This method enables differentiation through spikes, allowing our model fully end-to-end trainable.

%% file: sec/4_evaluation.tex
\section{Evaluation}

\subsection{Experimental Setup}

In this study, we explore a classification task that utilizes the CIFAR-100 dataset \citep{krizhevsky2009learning}. It comprises 60,000 images; each is a 32x32 pixel color image distributed across 100 categories, with 600 images in each category. This dataset is particularly suitable for deployment on edge devices since they often process low-resolution images.

For our analysis, we opted for the widely recognized ResNet50 and MobileNetV1 architectures. We identified each residual block (RB) within ResNet50 and each inverted residual block (IRB) within MobileNetV1 as potential points for architectural modification. The ResNet50 model comprises 16 RBs, while the MobileNetV1 architecture contains 13 IRBs, resulting in the creation of an equal number of distinct models. Each model is defined by strategically positioning the bottleneck unit after one of these blocks. This placement yields varying degrees of reduction in both the channel and spatial dimensions across the respective configurations. 

\begin{table*}[h!]
\centering
\begin{tabular}{cccccc}
\toprule
\toprule
\textbf{Architecture} & \textbf{Split Pt.} & \textbf{T-Step} & \textbf{Acc. (Drop)} &  \textbf{Bit Red.} & \textbf{Energy Imp.} \\
\midrule
ResNet50 & 4 & 2 & 73.31 (0.76)\% & 16x & 10x \\
&  & 4 & 75.16 (1.1)\% & 32x & 6x \\
\midrule
MobileNetv1 & 3 & 2 & 69.82 (1.21)\%  & 16x & 8x \\
&  & 4 & 73.37 (1.47) \% & 64x & 4x\\
\bottomrule
\end{tabular}
\caption{Impact of Time Steps on SpikeBottleNet Performance}
\label{table:ablation1}
\end{table*}

\begin{table}[h!]
\centering
\begin{tabular}{ccc}
\toprule
\toprule
\textbf{Model} & \textbf{Time Step} & \textbf{Accuracy (\%)} \\
\midrule
\multirow{2}{10em}{ResNet50} & 4 & 76.27 \\
 & 2 & 74.07 \\
 \midrule
\multirow{2}{10em}{MobileNetV1} & 4 & 74.86 \\
 & 2 & 71.03 \\
\bottomrule
\end{tabular}
\caption{Accuracy of the spike version of the models at different time steps on CIFAR100 dataset}
\label{table:spikenets-accuracy}
\end{table}

While our spike version of ResNet50 and MobileNetV1 may not surpass the current state-of-the-art performance on the CIFAR-100 dataset, the primary focus of this study lies in assessing the performance degradation, compression capabilities and energy efficiency achieved by our approach. The training phase accuracies of the models are reported in the Table \ref{table:spikenets-accuracy}. To maintain a high feature compression ratio, the time step should be limited to 4 or fewer, as additional time steps linearly increase latency and energy consumption. The cross-entropy function served as the loss function during the training process.

\subsection{Compression Capability Comparison}
Traditional DNNs based edge-cloud co-inference systems use continuous values for activations and feature compression during both training and inference phases \citep{Eshratifar2019}. These systems typically require high precision, often employing 32-bit floating-point (FP) representations, to preserve model accuracy. However, techniques like quantization \citep{kuzmin2024pruning} can reduce the bit-width requirements, typically from 32-bit floating-point (FP32) to lower bit-width formats like 16-bit (FP16), 8-bit (INT8), or even 4-bit. However, the most significant challenge with reducing bit width is the potential loss in model accuracy. Furthermore, lower precision formats can lead to rounding errors, reduced precision in gradient updates, and less representational capacity. Conversely, our SNN based SpikeBottleNet relies on spikes, which inherently lower the bit requirements for encoding neuron states. These spikes can be efficiently encoded using just 1-bit representations, which significantly reduces the communication overhead and bitstream size. In the following, we compare the compression capability of our SpikeBottleNet with other methods, such as BottleNet \citep{Eshratifar2019} \citep{Shao2020} across different splitting points. The BottleNet architecture used in this comparison is our implemented version.

\begin{figure}[ht]
\centering
\includegraphics[width=\linewidth, keepaspectratio]{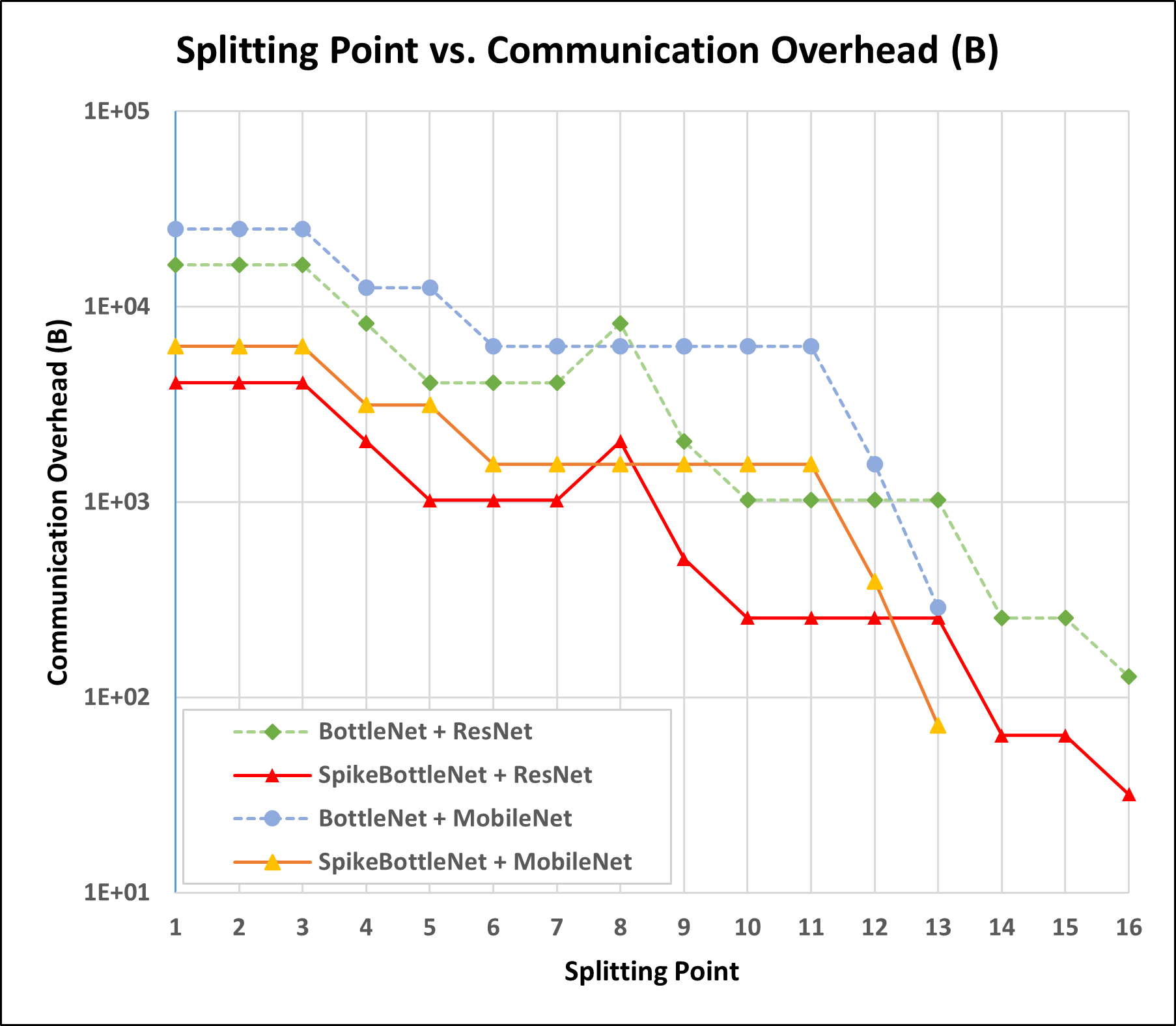}
\caption{Compression capability comparison between SpikeBottleNet and BottleNet}
\label{fig:compression-capability}
\end{figure}

The results depicted in Figure \ref{fig:compression-capability} reveal that, across all splitting points with a time step 2, SpikeBottleNet consistently achieves the lowest communication overhead compared to the DNN-based BottleNet. This efficiency is highlighted by its achievement of up to 256x compression ratio in the final convolutional layer of the SpikeBottleNet + ResNet50 architecture, compressing 65,536 bits (8 KB) down to 256 bits (32 Bytes), with an optimal accuracy reduction of only 0.16\%. In comparison, SpikeBottleNet demonstrates a 75\% reduction in the data size of 32 bytes required by the BottleNet + ResNet50 (128 bytes). This 75\% reduction in data size by SpikeBottleNet occurs consistently across all evaluated splitting points compared to the baseline. Similarly, the compression strategies for SpikeBottleNet + MobileNet effectively reduce data dimensions, achieving significant compression ratios, particularly at final layers (up to 174x), with minimal accuracy loss of 0.720\%. This highlights SpikeBottleNet's superior feature compression capability, enabling more efficient communication than traditional compression methods.

\subsection{Theoretical Energy Efficiency Estimation}
Our analysis of SpikeBottleNet's theoretical energy consumption is based on the established research \citep{hu2021advancing}\citep{hu2021spiking}. We assume an implementation on 45nm mobile hardware, where the energy cost per MAC operation is 4.6pJ and per AC operation is 0.9pJ \citep{horowitz20141}. Since SpikeBottleNet operates without requiring MAC operations and relies solely on AC, its significantly lower energy consumption is evident. Our calculated energy consumption represents the edge device's energy usage up to a specific splitting point. This analysis concentrates on the convolutional (CONV) layers within the residual blocks, which constitute a significant portion of the network's FLOPs. The computational cost from the edge device's encoder is excluded from analysis, as the encoder is a lightweight layer and its operational cost remains constant.

To assess the energy consumption of SpikeBottleNet, we first calculate the cumulative synaptic operations ($SyOPs$) for each block/layer ($l$) up to a certain splitting point using the following equation:

\begin{equation}
SyOPs(l)\ =\ fr\ \ast\ T\ \ast\ FLOPs(l)
\label{eq:evaluation1}
\end{equation}

where, $l$\ is the block or layer index in SpikeBottleNet, $fr$ represents firing rate of the input spike train reaching the block/layer, $T$ denotes simulation time step of the spiking neuron model, $FLOPs(l)$ indicate the number of floating-point operations up to layer $l$, representing the multiply-and-accumulate (MAC) operations, and $SyOPs$ signifies number of spike-based accumulate (AC) operations.

The firing rate $fr$ is defined as the probability of a neuron firing per timestep. It is estimated under a batch of random samples as follows: 

\begin{equation}
fr=\frac{1}{LN}\sum_{l=1}^{L}\sum_{i=1}^{N}o_{i,l}
\label{eq:evaluation2}
\end{equation}

where $L$ is the total number of layers until splitting points, $N$ is the number of neurons per layer, and $o_{i,l}$ represents the spike activity of neuron $i$ in layer $l$.

\begin{figure}[ht]
\centering
\includegraphics[width=\linewidth, keepaspectratio]{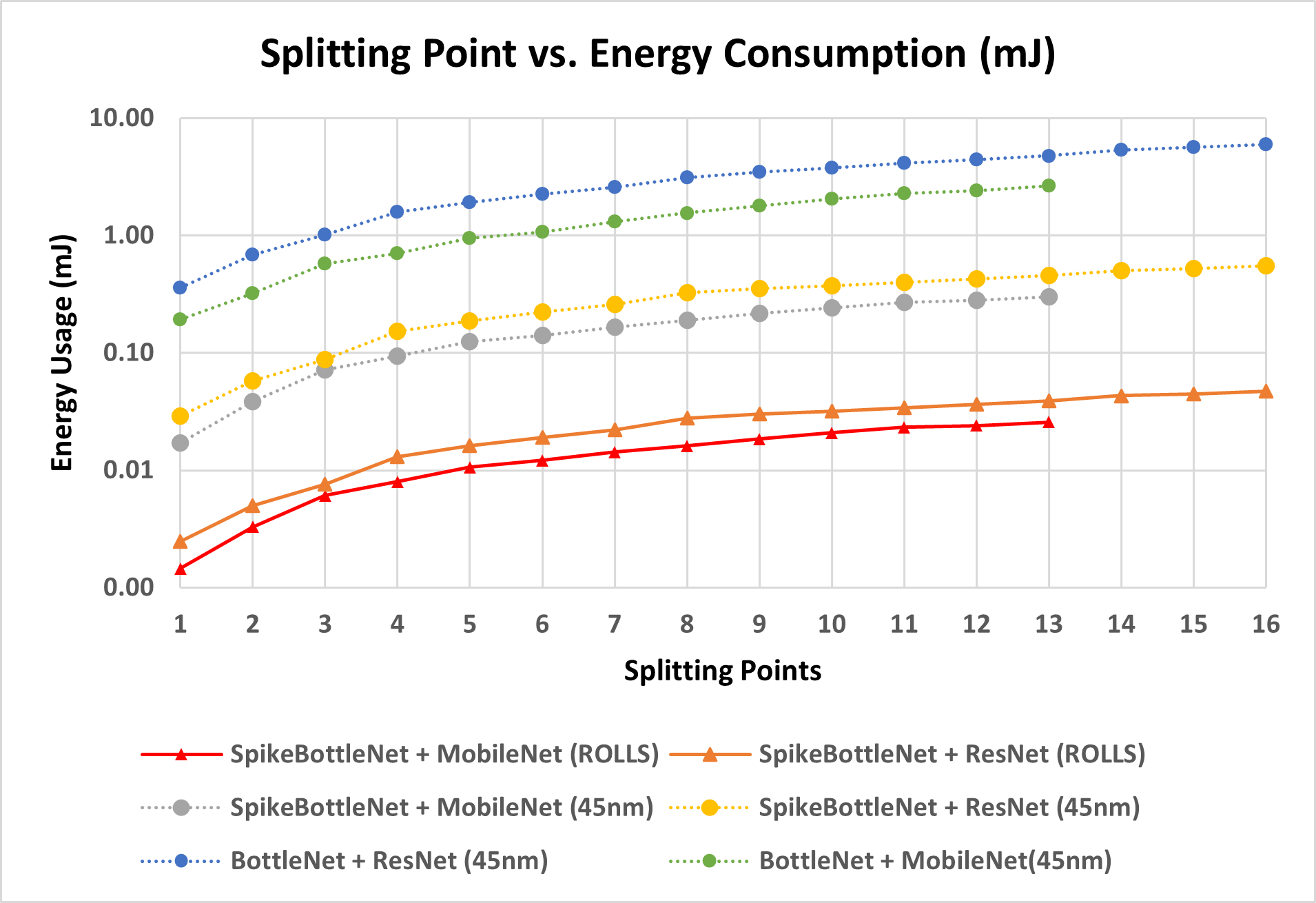}
\caption{Comparison of mobile energy consumption across various partition points for traditional 45nm hardware and ROLLS neuromorphic hardware}
\label{fig:energy-comparison}
\end{figure}

The results in Figure \ref{fig:energy-comparison} demonstrate that, on both traditional and neuromorphic hardware, SpikeBottleNet exhibits greater power efficiency compared to BottleNet across all splitting points. For instance, it achieves over 12x energy reduction on 45nm hardware at the first splitting point of the ResNet architecture. Similary, SpikeBottleNet offers a 10x less energy consumption compared to BottleNet at the first splitting point of MobileNetV1. This trend holds consistently across most of the splitting points.

To estimate energy consumption on neuromorphic chip, we considered the ROLLS neuromorphic chip. Fabricated using a 180nm CMOS process, the ROLLS chip reportedly consumes 77fJ per SyOPs \citep{indiveri2015neuromorphic}. In our experiments, SpikeBottleNet exhibits up to 144x greater power efficiency at first splitting points of ResNet on ROLLS hardware compare to BottleNet on conventional hardware. Detailed results are provided in Supplementary Materials Section 8. This finding underscores the promising future of SNN-based SpikeBottleNet. The estimated power consumption of SNNs implemented on neuromorphic hardware significantly surpasses that of DNN-based BottleNet.

\subsection{Ablation Study}
\subsubsection{Impact of Time Step Changes on SpikeBottleNet's Performance}
This ablation study investigates the impact of varying time steps within the SpikeBottleNet architecture on accuracy, bit reduction, and energy efficiency when deployed on conventional 45nm hardware. We selected one-fourth of the splitting points in each network for the ablation study. By dividing the network into four equal segments, we minimize computational demands on edge devices and reduce power consumption. Our experiments reported in Table \ref{table:ablation1} demonstrate that increasing time steps effectively enhances bit reduction but introduces a trade-off in energy efficiency. For example, with 4 time steps, ResNet50 achieves a 32x reduction in bitrate and a 6x improvement in energy consumption, while MobileNetv1 realizes a 64x bit reduction with a 4x energy gain. Detailed results are available in the supplementary materials Section 6 and Section 7.

While increasing time steps generally leads to improved accuracy and data compression, it is essential to carefully consider the potential trade-offs in energy efficiency and latency when deploying SpikeBottleNet in edge-cloud environments.

%% file: sec/5_conclusion.tex
\section{Conclusions}
This work presents a novel collaborative intelligence framework, SpikeBottleNet, that tackles the energy limitations hindering the deployment of deep neural networks on mobile devices. Our SNN partitioning strategy significantly reduces the communication cost of transferring features from the mobile device to the cloud, while minimizing the mobile device's energy consumption during co-inference at the edge-cloud. By exploiting the inherent binary spike signals employed by SNNs, SpikeBottleNet utilizes low-power accumulation methods instead of the high-power multiply-accumulate operations. This substitution leads to significantly minimizing the mobile device's energy footprint. Our evaluations demonstrate that the proposed SNN-based SpikeBottleNet achieves a remarkable 256x bit compression ratio in the final convolutional layer. Moreover, it translates to a 75\% reduction in bit transmission at most of the splitting points compared to the baseline model, with a measured accuracy degradation of only less than 2\%. Furthermore, SpikeBottleNet is well-suited for deployment on neuromorphic hardware, offering up to 144x energy savings on mobile devices.

Future work will focus on reducing SpikeBottleNet’s timesteps from multiple to a single step, aiming to further minimize latency, communication costs, and energy usage. Additional goals include exploring dynamic timesteps to potentially enhance accuracy, compression, and energy efficiency, optimizing the model for even more effective edge-cloud co-inference solution.

%% file: sec/X_suppl.tex
\clearpage
\setcounter{page}{1}
\maketitlesupplementary

\section{Impact Analysis of Compression Techniques on ResNet50}
\label{sec:compression-impact-resnet}

The Table \ref{table:resnet-accuracy-timestep1} and Table \ref{table:resnet-accuracy-timestep2} provide a detailed analysis of the impact of applying BottleNet and SpikeBottleNet compression techniques to ResNet50 at different network layers. The analysis considers feature dimensions, compressed feature dimensions, required byte count, compression ratios, and accuracy loss. As compression is applied to deeper layers, feature dimensions diminish, and compression ratios generally increase, reaching a maximum of 256x at layer 16. While higher compression ratios often result in substantial accuracy drops, as exemplified by the 1.89\% decrease at layer 8, some deeper layers, such as layer 16, achieve high compression (256x) with minimal accuracy loss (0.16\%). This analysis underscores the delicate balance between data reduction and model performance across various network depths.

\section{Impact Analysis of Compression Techniques on MobileNetV1}
\label{sec:compression-impact-mobilenet}
Table \ref{table:mobilenet-accuracy-timestep1} and \ref{table:mobilenet-accuracy-timestep2} present the performance of splitting SpikeBottleNet at different layers of MobileNetV1. Early layer splits (1-3) result in significant data size reductions (25,088 bytes) with moderate compression ratios (16-32x). Notably, these early splits incur only minor accuracy drops (~1.3\%). As the splitting point moves deeper into the network (layers 12-13), the compression ratios increase dramatically (up to 174x), reducing data sizes to as low as 288 bytes. However, the impact on accuracy becomes more pronounced, with drops ranging from 0.72\% to 2.04\%. This analysis underscores a clear trade-off between compression and accuracy. Deeper splits offer higher compression but may compromise model performance, whereas earlier splits provide a more balanced approach, trading off some compression for improved accuracy.

\begin{table*}[h]
\begin{tabular*}{\textwidth}{@{\extracolsep{\fill}}p{3cm}p{1.2cm}p{1.2cm}p{1.2cm}p{1.2cm}p{1.2cm}p{1.2cm}p{1.2cm}p{1.2cm}}
\toprule
\toprule
\textbf{Splitting Point} & \textbf{1} & \textbf{2} & \textbf{3} & \textbf{4} & \textbf{5} & \textbf{6} & \textbf{7} & \textbf{8} \\
\midrule \textbf{Time Step} & 2 & 2 & 2 & 2 & 2 & 2 & 2 & 2 \\
\midrule \textbf{Original} & 256x32x32 & 256x32x32 & 256x32x32 & 512x16x16 & 512x16x16 & 512x16x16 & 512x16x16 & 1024x8x8 \\
\midrule \textbf{Compressed} & 16x32x32 & 16x32x32 & 16x32x32 & 32x16x16 & 16x16x16 & 16x16x16 & 16x16x16 & 128x8x8 \\
\midrule \textbf{BottleNet (B)} & 16,384 & 16,384 & 16,384 & 8,192 & 4,096 & 4,096 & 4,096 & 8,192 \\
\midrule \textbf{SpikeBottleNet (B)} & 4,096 & 4,096 & 4,096 & 2,048 & 1,024 & 1,024 & 1,024 & 2,048 \\
\midrule \textbf{Compression Ratio} & 16 & 16 & 16 & 16 & 32 & 32 & 32 & 8 \\
\midrule \textbf{Accuracy Drop (\%)} & 0.67 & 0.27 & 0.09 & 0.36 & 0.98 & 0.36 & 0.21 & 1.89 \\ \bottomrule
\bottomrule
\end{tabular*}
\caption{Compression and Accuracy Trade-offs Across Different Splitting Points  on ResNet50 architecture (1 - 8) in SpikeBottleNet}
\label{table:resnet-accuracy-timestep1}
\end{table*}

\begin{table*}[h]
\begin{tabular*}{\textwidth}{@{\extracolsep{\fill}}p{3cm}p{1.2cm}p{1.2cm}p{1.2cm}p{1.2cm}p{1.2cm}p{1.2cm}p{1.2cm}p{1.2cm}}
\toprule
\midrule \textbf{Splitting Point} & \textbf{9} & \textbf{10} & \textbf{11} & \textbf{12} & \textbf{13} & \textbf{14} & \textbf{15} & \textbf{16} \\
\midrule \textbf{Time Step} & 2 & 2 & 2 & 2 & 2 & 2 & 2 & 2 \\
\midrule \textbf{Original} & 1024x8x8 & 1024x8x8 & 1024x8x8 & 1024x8x8 & 1024x8x8 & 2048x4x4 & 2048x4x4 & 2048x4x4 \\
\midrule \textbf{Compressed} & 32x8x8 & 16x8x8 & 16x8x8 & 16x8x8 & 16x8x8 & 16x4x4 & 16x4x4 & 8x4x4 \\
\midrule \textbf{BottleNet (B)} & 2,048 & 1,024 & 1,024 & 1,024 & 1,024 & 256 & 256 & 128 \\
\midrule \textbf{SpikeBottleNet (B)} & 512 & 256 & 256 & 256 & 256 & 64 & 64 & 32 \\
\midrule \textbf{Compression Ratio} & 32 & 64 & 64 & 64 & 64 & 128 & 128 & 256 \\
\midrule \textbf{Accuracy Drop (\%)} & 0.83 & 1.60 & 1.36 & 1.23 & 1.06 & 1.53 & 1.22 & 0.16 \\
\bottomrule
\bottomrule
\end{tabular*}
\caption{Extended Compression and Accuracy Trade-offs Across Different Splitting Points on ResNet50 architecture (8 - 16) in SpikeBottleNet}
\label{table:resnet-accuracy-timestep2}
\end{table*}

\begin{table*}[h!]
\centering
\begin{tabular*}{\textwidth}{@{\extracolsep{\fill}}p{3cm}p{1.2cm}p{1.2cm}p{1.2cm}p{1.2cm}p{1.2cm}p{1.2cm}p{1.2cm}}
\toprule
\midrule \textbf{Splitting Point} & \textbf{1} & \textbf{2} & \textbf{3} & \textbf{4} & \textbf{5} & \textbf{6} & \textbf{7} \\
\midrule
\textbf{Time Step} & 2 & 2 & 2 & 2 & 2 & 2 & 2 \\
\midrule \textbf{Original} & 64x112x112 & 128x56x56 & 128x56x56 & 256x28x28 & 256x28x28 & 512x14x14 & 512x14x14 \\
\midrule \textbf{Compressed} & 8x56x56 & 8x56x56 & 8x56x56 & 16x28x28 & 16x28x28 & 32x14x14 & 32x14x14 \\
\midrule \textbf{BottleNet (B)} & 25,088 & 25,088 & 25,088 & 12,544 & 12,544 & 6,272 & 6,272 \\
\midrule \textbf{SpikeBottleNet (B)} & 6,272 & 6,272 & 6,272 & 3,136 & 3,136 & 1,568 & 1,568 \\
\midrule \textbf{Compression Ratio} & 32 & 16 & 16 & 16 & 16 & 16 & 16 \\
\midrule \textbf{Accuracy Drop} (\%) & 1.34 & 1.35 & 1.21 & 1.47 & 0.78 & 0.77 & 0.73 \\
\bottomrule
\bottomrule
\end{tabular*}
\caption{Compression and Accuracy Trade-offs Across Different Splitting Points on MobileNetV1 architecture (1 - 7) in SpikeBottleNet}
\label{table:mobilenet-accuracy-timestep1}
\end{table*}

\begin{table*}[h!]
\centering
\begin{tabular*}{\textwidth}{@{\extracolsep{\fill}}p{3cm}p{1.2cm}p{1.2cm}p{1.2cm}p{1.2cm}p{1.2cm}p{1.2cm}}
\toprule
\textbf{Splitting Point} & \textbf{8} & \textbf{9} & \textbf{10} & \textbf{11} & \textbf{12} & \textbf{13} \\
\midrule
\textbf{Time Step} & 2 & 2 & 2 & 2 & 2 & 2 \\
\midrule
\textbf{Original} & 512x14x14 & 512x14x14 & 512x14x14 & 512x14x14 & 1024x7x7 & 1024x7x7 \\
\midrule
\textbf{Compressed} & 32x14x14 & 32x14x14 & 32x14x14 & 32x14x14 & 32x7x7 & 32x3x3 \\
\midrule
\textbf{BottleNet (B)} & 6,272 & 6,272 & 6,272 & 6,272 & 1,568 & 288 \\
\midrule
\textbf{SpikeBottleNet (B)} & 1,568 & 1,568 & 1,568 & 1,568 & 392 & 72 \\
\midrule
\textbf{Compression Ratio} & 16 & 16 & 16 & 16 & 32 & 174 \\
\midrule
\textbf{Accuracy Drop (\%)} & 0.58 & 0.52 & 0.84 & 1.16 & 2.04 & 0.72 \\
\bottomrule
\bottomrule
\end{tabular*}
\caption{Extended Compression and Accuracy Trade-offs Across Different Splitting Points on MobileNetV1 architecture (8 - 13) in SpikeBottleNet}
\label{table:mobilenet-accuracy-timestep2}
\end{table*}

\section{Impact Analysis of Energy Efficiency of SpikeBottleNet and BottleNet Architectures}
\label{sec:energy-impact}
The provided Tables \ref{table:resnet-energy-1} - \ref{table:mobilenet-energy-2}  compare the energy consumption of SpikeBottleNet and BottleNet architectures when applied to ResNet50 and MobileNetV1 models at various splitting points. For both architectures, SpikeBottleNet consistently outperforms BottleNet in terms of energy efficiency.

For ResNet50, the energy efficiency ratio stabilizes at approximately 10-11 for the 45nm chip and 120-130 for the ROLLS chip at higher splitting points. In contrast, MobileNetV1 displays slightly lower stabilization values, ranging from 7-9 for the 45nm chip and 88-102 for the ROLLS chip. This suggests that SpikeBottleNet's energy efficiency advantage is more pronounced in MobileNetV1, particularly at lower splitting points. 

% First Table (Splitting Points 1 to 8)
\begin{table*}[h]
\centering
\begin{tabular}{lcccccccc}
\toprule
\toprule
\multicolumn{9}{c}{\textbf{32 bit-FP: MAC 4.6pJ, AC 0.9 pJ}} \\
\midrule
\textbf{Splitting Point} & 1 & 2 & 3 & 4 & 5 & 6 & 7 & 8 \\
\midrule \textbf{GFLOPs} & 0.08 & 0.15 & 0.22 & 0.34 & 0.41 & 0.49 & 0.56 & 0.68 \\
\midrule
\textbf{BottleNet (mJ)} & 0.36 & 0.69 & 1.02 & 1.58 & 1.91 & 2.24 & 2.57 & 3.13 \\
\midrule \textbf{SpikeBottleNet 45nm (mJ)} & 0.03 & 0.06 & 0.09 & 0.15 & 0.19 & 0.22 & 0.26 & 0.33 \\
\midrule \textbf{$\frac{E(\text { BottleNet })}{E(\text { SpikeBottleNet })}$} & 12.37 & 11.88 & 11.56 & 10.35 & 10.16 & 10.05 & 9.91 & 9.59 \\
\midrule
\multicolumn{9}{c}{\textbf{ROLLS chip: 77fJ per SyOPs}} \\
\midrule \textbf{Firing Rate} & 0.2066 & 0.2151 & 0.2210 & 0.2468 & 0.2515 & 0.2542 & 0.2579 & 0.2664 \\
\midrule \textbf{GSyOPs} & 0.03 & 0.06 & 0.10 & 0.17 & 0.21 & 0.25 & 0.29 & 0.36 \\
\midrule
\textbf{SpikeBottleNet ROLLS (mJ)} & 0.002 & 0.005 & 0.008 & 0.013 & 0.016 & 0.019 & 0.022 & 0.028 \\
\midrule \textbf{$\frac{E(\text { BottleNet })}{E(\text { SpikeBottleNet })}$} & 144.58 & 138.87 & 135.16 & 121.03 & 118.77 & 117.51 & 115.82 & 112.13 \\
\bottomrule
\bottomrule
\end{tabular}
\caption{Energy Consumption Comparison for Splitting Points 1 to 8 of ResNet50}
\label{table:resnet-energy-1}
\end{table*}

% Second Table (Splitting Points 9 to 16)
\begin{table*}[h]
\centering
\begin{tabular}{lcccccccc}
\toprule
\toprule
\multicolumn{9}{c}{\textbf{32 bit-FP: MAC 4.6pJ, AC 0.9 pJ}} \\
\midrule
\textbf{Splitting Point} & 9 & 10 & 11 & 12 & 13 & 14 & 15 & 16 \\
\midrule
\textbf{GFLOPs} & 0.75 & 0.82 & 0.89 & 0.97 & 1.04 & 1.16 & 1.23 & 1.30 \\
\midrule
\textbf{BottleNet (mJ)} & 3.46 & 3.78 & 4.11 & 4.44 & 4.77 & 5.33 & 5.66 & 5.99 \\
\midrule \textbf{SpikeBottleNet 45nm (mJ)} & 0.35 & 0.37 & 0.40 & 0.42 & 0.46 & 0.50 & 0.52 & 0.55 \\
\midrule \textbf{$\frac{E(\text { BottleNet })}{E(\text { SpikeBottleNet })}$} & 9.83 & 10.17 & 10.34 & 10.47 & 10.47 & 10.58 & 10.87 & 10.87 \\
\midrule
\multicolumn{8}{c}{\textbf{ROLLS chip: 77fJ per SyOPs}} \\
\midrule \textbf{Firing Rate} & 0.2599 & 0.2513 & 0.2471 & 0.2442 & 0.2440 & 0.2416 & 0.2350 & 0.2352 \\
\midrule \textbf{GSyOPs} & 0.39 & 0.41 & 0.44 & 0.47 & 0.51 & 0.56 & 0.58 & 0.61 \\
\midrule
\textbf{SpikeBottleNet ROLLS (mJ)} & 0.030 & 0.032 & 0.034 & 0.036 & 0.039 & 0.043 & 0.045 & 0.047 \\
\midrule \textbf{$\frac{E(\text { BottleNet })}{E(\text { SpikeBottleNet })}$} & 114.93 & 118.86 & 120.88 & 122.32 & 122.42 & 123.63 & 127.11 & 127.00 \\
\bottomrule
\bottomrule
\end{tabular}
\caption{ Extended Energy Consumption Comparison for Splitting Points 9 to 16 of ResNet50}
\label{table:resnet-energy-2}
\end{table*}

% First Table (Splitting Points 1 to 7)
\begin{table*}[h]
\centering
\begin{tabular}{lcccccccc}
\toprule
\toprule
\multicolumn{8}{c}{\textbf{32 bit-FP: MAC 4.6pJ, AC 0.9 pJ}} \\
\midrule
\textbf{Splitting Point} & 1 & 2 & 3 & 4 & 5 & 6 & 7 \\
\midrule
\textbf{GFLOPs} & 0.04 & 0.07 & 0.13 & 0.15 & 0.21 & 0.23 & 0.28 \\
\midrule \textbf{BottleNet (mJ)} & 0.19 & 0.32 & 0.58 & 0.70 & 0.95 & 1.07 & 1.31 \\
\midrule \textbf{SpikeBottleNet 45nm (mJ)} & 0.02 & 0.04 & 0.07 & 0.09 & 0.12 & 0.14 & 0.17 \\
\midrule \textbf{E(BottleNet) / E(SpikeBottleNet)} & 11.28 & 8.35 & 8.12 & 7.53 & 7.66 & 7.55 & 7.88 \\
\midrule
\multicolumn{8}{c}{\textbf{ROLLS chip: 77fJ per SyOPs}} \\
\midrule
\midrule \textbf{GSyOPs} & 0.02 & 0.04 & 0.08 & 0.10 & 0.14 & 0.16 & 0.18 \\
\midrule \textbf{Firing Rate} & 0.2265 & 0.3059 & 0.3146 & 0.3396 & 0.3337 & 0.3384 & 0.3242 \\
\midrule \textbf{SpikeBottleNet ROLLS (mJ)} & 0.001 & 0.003 & 0.006 & 0.008 & 0.011 & 0.012 & 0.014 \\
\midrule \textbf{$\frac{E(\text { BottleNet })}{E(\text { SpikeBottleNet })}$} & 131.88 & 97.65 & 94.95 & 87.96 & 89.51 & 88.27 & 92.13 \\
\bottomrule
\bottomrule
\end{tabular}
\caption{Energy Consumption Comparison for Splitting Points 1 to 7 of MobileNetV1}
\label{table:mobilenet-energy-1}
\end{table*}

% Second Table (Splitting Points 8 to 13)
\begin{table*}[h]
\centering
\begin{tabular}{lcccccc}
\toprule
\toprule
\multicolumn{7}{c}{\textbf{32 bit-FP: MAC 4.6pJ, AC 0.9 pJ}} \\
\midrule
\textbf{Splitting Point} & 8 & 9 & 10 & 11 & 12 & 13 \\
\midrule
\textbf{GFLOPs} & 0.34 & 0.39 & 0.44 & 0.49 & 0.52 & 0.57 \\
\midrule \textbf{BottleNet (mJ)} & 1.55 & 1.79 & 2.03 & 2.28 & 2.40 & 2.63 \\
\midrule \textbf{SpikeBottleNet 45nm (mJ)} & 0.19 & 0.22 & 0.24 & 0.27 & 0.28 & 0.30 \\
\midrule \textbf \textbf{$\frac{E(\text { BottleNet })}{E(\text { SpikeBottleNet })}$} & 8.23 & 8.30 & 8.37 & 8.41 & 8.53 & 8.76 \\
\midrule
\multicolumn{7}{c}{\textbf{ROLLS chip: 77fJ per SyOPs}} \\
\midrule
\textbf{GSyOPs} & 0.21 & 0.24 & 0.27 & 0.30 & 0.31 & 0.33 \\
\midrule \textbf{Firing Rate} & 0.3105 & 0.3078 & 0.3054 & 0.3038 & 0.2997 & 0.2918 \\
\midrule \textbf{SpikeBottleNet ROLLS (mJ)} & 0.016 & 0.018 & 0.021 & 0.023 & 0.024 & 0.026 \\
\midrule \textbf \textbf{$\frac{E(\text { BottleNet })}{E(\text { SpikeBottleNet })}$} & 96.20 & 97.04 & 97.81 & 98.32 & 99.67 & 102.37 \\
\bottomrule
\bottomrule
\end{tabular}
\caption{Energy Consumption Comparison for Splitting Points 8 to 13 of MobileNetV1}
\label{table:mobilenet-energy-2}
\end{table*}